\pdfoutput=1
\documentclass[11pt,a4paper]{article}
\usepackage[T1]{fontenc}
\usepackage[hyperref]{udst2018}
\usepackage{times}
\usepackage{latexsym}
\usepackage{amsmath}
\usepackage{tikz}
\usepackage{tikz-dependency}
\usepackage[warn]{textcomp}
\usepackage[font=small]{caption}
\usepackage{subcaption}
\usepackage{cprotect}
\usepackage{multirow}
\usepackage{url}
\usepackage{etoolbox}
\usepackage{xr}
\usepackage{adjustbox}
\usepackage{pgfplots}
\usepackage{pgfplotstable}

\newcommand{\com}[1]{}

\pgfplotsset{select coords between index/.style 2 args={
    x filter/.code={
        \ifnum\coordindex<#1\fi
        \ifnum\coordindex>#2\fi
    }
}}

\makeatletter
\patchcmd\@combinedblfloats{\box\@outputbox}{\unvbox\@outputbox}{}{%
   \errmessage{\noexpand\@combinedblfloats could not be patched}%
}%
 \makeatother

\usetikzlibrary{shapes,shapes.misc}

\aclfinalcopy 
\def\aclpaperid{8} %  Enter the Paper ID here for final camera ready copy

\title{Universal Dependency Parsing with a \\ General Transition-Based DAG Parser}

\author{Daniel Hershcovich$^{1,2}$ \\
  \\\And
  Omri Abend$^2$ \\
  $^1$The Edmond and Lily Safra Center for Brain Sciences \\
  $^2$School of Computer Science and Engineering \\
  Hebrew University of Jerusalem \\
  \texttt{\{danielh,oabend,arir\}@cs.huji.ac.il}
  \\\And
  Ari Rappoport$^2$
}

\date{}

\begin{document}
\maketitle
\begin{abstract}
This paper presents our experiments with applying TUPA to the CoNLL 2018 UD shared task. TUPA is a general neural transition-based DAG parser, which we use to present the first experiments on recovering enhanced dependencies as part of the general parsing task. TUPA was designed for parsing UCCA, a cross-linguistic semantic annotation scheme, exhibiting reentrancy, discontinuity and non-terminal nodes. By converting UD trees and graphs to a UCCA-like DAG format, we train TUPA almost without modification on the UD parsing task. The generic nature of our approach lends itself naturally to multitask learning.  
Our code is available at \url{https://github.com/CoNLL-UD-2018/HUJI}.
\end{abstract}

\section{Introduction}\label{sec:introduction}

In this paper, we present the HUJI submission to the CoNLL 2018 shared task
on Universal Dependency parsing \cite{udst:overview2018}.
We focus only on parsing, using the baseline system, UDPipe 1.2
\cite{udpipe,udpipe:2017}
for tokenization, sentence splitting, part-of-speech tagging and morphological tagging.

Our system is based on TUPA \cite[see \S\ref{sec:model}]{hershcovich2017a,hershcovich2018multitask},
a transition-based UCCA parser.
UCCA \cite[Universal Conceptual Cognitive Annotation;][]{abend2013universal} is a
cross-linguistic semantic annotation scheme, representing events, participants,
attributes and relations in a directed acyclic graph (DAG) structure.
UCCA allows reentrancy to support argument sharing,
discontinuity (corresponding to non-projectivity in dependency formalisms)
and non-terminal nodes (as opposed to dependencies, which are bi-lexical).
To parse Universal Dependencies \cite{nivre2016universal}
using TUPA, we employ a bidirectional conversion protocol to represent
UD trees and graphs in a UCCA-like unified DAG format (\S\ref{sec:format}).

\paragraph{Enhanced dependencies.}
Our method treats \textit{enhanced dependencies}\footnote{\url{http://universaldependencies.org/u/overview/enhanced-syntax.html}}
as part of the dependency graph, providing the first approach, to our knowledge,
for supervised learning of enhanced UD parsing.
Due to the scarcity of enhanced dependencies in UD treebanks,
previous approaches \cite{SCHUSTER16.779,D17-1009} have attempted to recover them
using language-specific rules.
Our approach attempts to learn them from data:
while only a few UD treebanks contain any enhanced dependencies,
similar structures are an integral part of UCCA and its annotated corpora
(realized as reentrancy by remote edges; see \S\ref{sec:format}),
and TUPA supports them as a standard feature.

As their annotation in UD is not yet widespread and standardized,
enhanced dependencies are \textit{not included}
in the evaluation metrics for UD parsing,
and so TUPA's ability to parse them is not reflected in the official
shared task scores.
However, we believe these enhancements, representing
case information, elided predicates,
and shared arguments due to conjunction, control, raising
and relative clauses,
provide richer information to downstream semantic applications,
making UD better suited for text understanding.
We propose an evaluation metric specific to enhanced dependencies,
\textit{enhanced~LAS} (\S\ref{sec:enhanced_results}),
and use it to evaluate our method.

%%%%%%%%%%%%%%%%%%%%%%%%%%%%%%%%%%%%%%%%%%%%%%%%%%%%%%%%%%%%%%%%%%%%%%%%%%%%%%%%%%%%%%%%

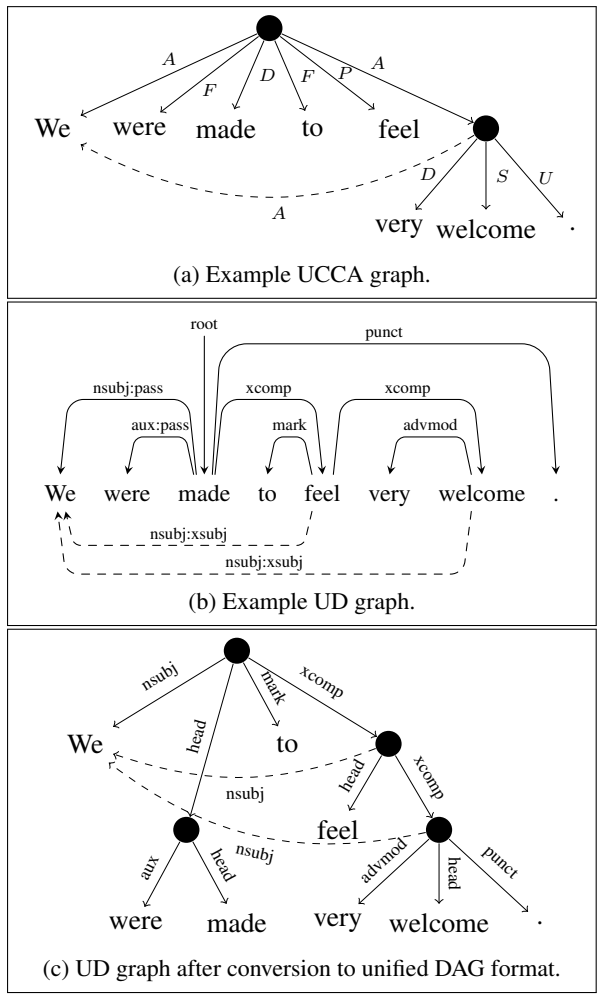
\begin{figure}[ht]
\fbox{\begin{subfigure}{0.47\textwidth}
    \centering
    \scalebox{.95}{
    \begin{tikzpicture}[level distance=14mm,sibling distance=12mm, ->,
        every circle node/.append style={fill=black}]
      \tikzstyle{word} = [font=\rmfamily,color=black]
      \node (ROOT) [circle] {}
        child {node (We) [word] {We} edge from parent node[above] {\scriptsize  $A$}}
        child {node [word] {were} edge from parent node[below] {\scriptsize $F$}}
        child {node [word] {made} edge from parent node[right] {\scriptsize $D$}}
        child {node [word] {to} edge from parent node[right] {\scriptsize $F$}}
        child {node [word] {feel} edge from parent node[right] {\scriptsize $P$}}
        child {node (verywelcome) [circle] {}
        {
          child {node [word] {very} edge from parent node[left] {\scriptsize $D$}}
          child {node [word] {welcome} edge from parent node[right] {\scriptsize $S$}}
          child {node [word] {.} edge from parent node[right] {\scriptsize $U$}}
        } edge from parent node[above] {\scriptsize $A$} }
      ;
      \draw[dashed,->,bend left] (verywelcome) to node [below] {\scriptsize $A$} (We);
    \end{tikzpicture}}
  \caption{Example UCCA graph.}
  \label{fig:converted_example_ucca}
\end{subfigure}}
\fbox{\begin{subfigure}{0.47\textwidth}
  \centering
    \begin{dependency}[text only label, label style={above}, font=\small]
    \begin{deptext}[column sep=.5em,ampersand replacement=\^]
    We \^ were \^ made \^ to \^ feel \^ very \^ welcome \^ . \\
    \end{deptext}
        \depedge[edge start x offset=1pt]{3}{1}{nsubj:pass}
        \depedge[edge below,dashed,edge unit distance=.65ex,edge end x offset=2pt]{5}{1}{nsubj:xsubj}
        \depedge[edge below,dashed,edge unit distance=.8ex,edge end x offset=-1pt]{7}{1}{nsubj:xsubj}
        \depedge{3}{2}{aux:pass}
        \deproot[edge unit distance=3ex]{3}{root}
        \depedge{5}{4}{mark}
        \depedge{3}{5}{xcomp}
        \depedge{7}{6}{advmod}
        \depedge{5}{7}{xcomp}
        \depedge[edge unit distance=2ex,edge start x offset=-1pt]{3}{8}{punct}
    \end{dependency}
  \caption{\label{fig:original_examples}Example UD graph.}
\end{subfigure}}
\fbox{\begin{subfigure}{0.47\textwidth}
  \centering
  \scalebox{.95}{
    \begin{tikzpicture}[sibling distance=14mm, ->,
        every circle node/.append style={fill=black},
        level 1/.style={level distance=13mm},
        level 2/.style={level distance=12mm},
        level 3/.style={level distance=13mm}]
      \tikzstyle{word} = [font=\rmfamily,color=black]
      \node (ROOT) [circle] {}
        child {node (We) [word] {We} edge from parent node[midway,above,sloped] {\scriptsize nsubj}}
        child {node {}
        {
	        child {node (weremade) [circle] {}
	        {
	          child {node [word] {were} edge from parent node[midway,above,sloped] {\scriptsize aux}}
	          child {node [word] {made} edge from parent node[midway,above,sloped] {\scriptsize head}}
	        } edge from parent [draw=none]}
        } edge from parent [draw=none]}
        child {node [word] {to} edge from parent node[midway,above,sloped] {\scriptsize mark}}
        child {node (feelverywelcome) [circle] {}
        {
          child {node [word] {feel} edge from parent node[midway,above,sloped] {\scriptsize head}}
          child {node (verywelcome) [circle] {}
          {
            child {node [word] {very} edge from parent node[midway,above,sloped] {\scriptsize advmod}}
            child {node [word] {welcome} edge from parent node[midway,above,sloped] {\scriptsize head}}
            child {node [word] {.} edge from parent node[midway,above,sloped] {\scriptsize punct}}
          } edge from parent node[midway,above,sloped] {\scriptsize xcomp} }
        } edge from parent node[midway,above,sloped] {\scriptsize xcomp} }
      ;
      \draw[->] (ROOT) to node [midway,above,sloped] {\scriptsize head} (weremade);
      \draw[dashed,->,bend left=25] (verywelcome) to node [midway,below,sloped] {\scriptsize nsubj} (We);
      \draw[dashed,->,bend left=20] (feelverywelcome) to node [midway,below,sloped] {\scriptsize nsubj} (We);
    \end{tikzpicture}}
  \captionof{figure}{UD graph after conversion to unified DAG format.}\label{fig:converted_example_ud}
\end{subfigure}}

\cprotect\caption{(a) Example UCCA annotation for the sentence
``We were made to feel very welcome.'',
containing a control verb, \textit{made}.
The dashed \textit{A} edge is a \textit{remote edge}.
(b) Bilexical graph annotating the same sentence in UD
(\verb|reviews-077034-0002| from \verb|UD_English-EWT|).
Enhanced dependencies appear below the sentence.
(c) The same UD graph, after conversion to the unified DAG format.
Intermediate non-terminals and \textit{head} edges are introduced,
to get a UCCA-like structure.}\label{fig:converted_examples}
\end{figure}

\begin{figure*}[!t]
\begin{adjustbox}{width=\textwidth,margin=3pt,frame}
\texttt{1 We we PRON PRP Case=Nom|Number=Plur|Person=1|PronType=Prs 3 nsubj:pass 3:nsubj:pass|5:nsubj:xsubj|7:nsubj:xsubj \textunderscore}
\end{adjustbox}
\cprotect\caption{Example line from CoNLL-U file with two enhanced dependencies:
\verb|5:nsubj:xsubj| and \verb|7:nsubj:xsubj|.}\label{fig:enhanced_conllu}
\end{figure*}

\begin{figure*}[ht]
\fbox{
  \centering
    \begin{dependency}[text only label, label style={above}, font=\small]
    \begin{deptext}[column sep=1.15em,ampersand replacement=\^]
    he \^ went \^ straight \^ to \^ work \^ and \^ finished \^ the \^ job \^ efficiently \^ and \^ promptly \^ ! \\
    \end{deptext}
        \depedge[edge start x offset=1pt]{2}{1}{nsubj}
        \depedge[edge below,dashed,edge unit distance=.5ex,edge end x offset=2pt]{7}{1}{nsubj}
        \deproot[edge unit distance=3ex]{2}{root}
        \depedge[edge start x offset=3pt]{2}{3}{advmod}
        \depedge{5}{4}{mark}
        \depedge[edge unit distance=1.75ex,edge start x offset=2pt]{2}{5}{advcl}
        \depedge{7}{6}{cc}
        \depedge[edge unit distance=1.5ex,edge start x offset=1pt]{2}{7}{conj}
        \depedge{9}{8}{det}
        \depedge[edge unit distance=2.5ex,edge start x offset=1pt]{7}{9}{obj}
        \depedge[edge unit distance=2.5ex,edge start x offset=-1pt]{7}{10}{advmod}
        \depedge{12}{11}{cc}
        \depedge{10}{12}{conj}
        \depedge[edge below,dashed,edge unit distance=.65ex,edge end x offset=2pt]{7}{12}{advmod}
        \depedge[edge unit distance=1ex,edge start x offset=-1pt]{2}{13}{punct}
    \end{dependency}
}
\cprotect\caption{UD graph from \verb|reviews-341397-0003| (\verb|UD_English-EWT|),
containing conjoined predicates and arguments.}
\label{fig:example_conj}
\end{figure*}

\begin{figure*}[ht]
\fbox{
  \centering
    \begin{dependency}[text only label, label style={above}, font=\small]
    \begin{deptext}[column sep=1.25em,ampersand replacement=\^]
    I \^ wish \^ all \^ happy \^ holidays \^ , \^ and \^ moreso \^ , \^ \textbf{E9.1} \^ peace \^ on \^ earth \^ . \\
    \end{deptext}
        \depedge[edge start x offset=1pt]{2}{1}{nsubj}
        \deproot[edge start x offset=-1pt,edge unit distance=3ex]{2}{root}
        \depedge[edge start x offset=3pt]{2}{3}{iobj}
        \depedge{5}{4}{amod}
        \depedge[edge start x offset=2pt]{2}{5}{obj}
        \depedge[edge unit distance=2ex]{11}{6}{punct}
        \depedge[edge start x offset=4pt,edge below,dashed]{10}{6}{punct}
        \depedge[edge start x offset=-1pt,edge unit distance=1.85ex]{11}{7}{cc}
        \depedge[edge start x offset=1pt,edge below,dashed]{10}{7}{cc}
        \depedge[edge start x offset=-3pt,edge unit distance=1.5ex]{11}{8}{orphan}
        \depedge[edge start x offset=-1pt,edge below,dashed]{10}{8}{advmod}
        \depedge[edge start x offset=-5pt,edge unit distance=.8ex]{11}{9}{punct}
        \depedge[edge start x offset=-3pt,edge below,dashed]{10}{9}{punct}
        \depedge[edge unit distance=1.45ex]{2}{11}{conj}
        \depedge[edge below,dashed]{10}{11}{obj}
        \depedge{13}{12}{case}
        \depedge[edge start x offset=1pt]{11}{13}{nmod}
        \depedge[edge start x offset=-1pt,edge unit distance=1.35ex]{2}{14}{punct}
    \end{dependency}
}
\cprotect\caption{\verb|newsgroup-groups.google.com_GuildWars_086f0f64ab633ab3_ENG_20041111_173500-0051| (\verb|UD_English-EWT|), containing a null node (\textbf{E9.1}) and
case information (nmod:on).}
\label{fig:example_ellipsis}
\end{figure*}

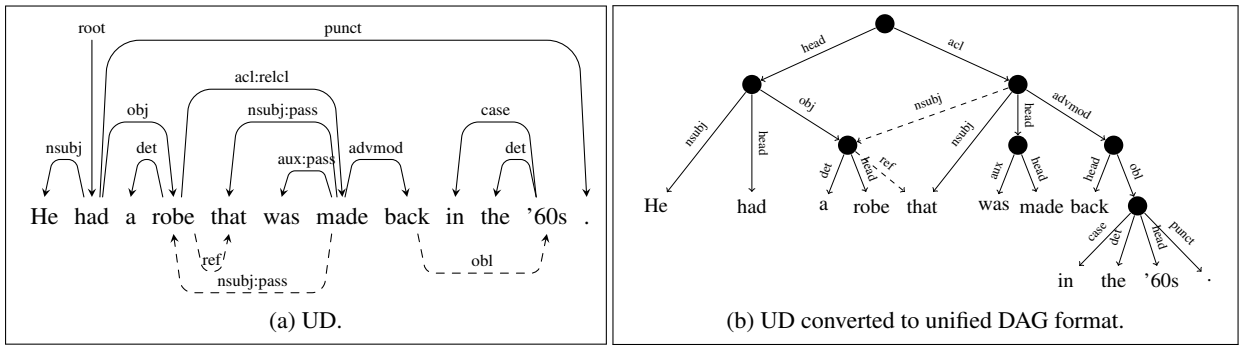
\begin{figure*}[ht]
\fbox{\begin{subfigure}{0.48\textwidth}
    \centering
    \begin{dependency}[text only label, label style={above}, font=\small]
    \begin{deptext}[column sep=.1em,ampersand replacement=\^]
    He \^ had \^ a \^ robe \^ that \^ was \^ made \^ back \^ in \^ the \^ '60s \^ . \\
    \end{deptext}
        \depedge[edge start x offset=1pt]{2}{1}{nsubj}
        \deproot[edge start x offset=-1pt,edge unit distance=3.355ex]{2}{root}
        \depedge{4}{3}{det}
        \depedge{2}{4}{obj}
        \depedge[edge unit distance=1.5ex,edge below,dashed]{7}{4}{nsubj:pass}
        \depedge[edge start x offset=2pt]{7}{5}{nsubj:pass}
        \depedge[edge start x offset=4pt,edge below,dashed]{4}{5}{ref}
        \depedge[edge unit distance=2ex]{7}{6}{aux:pass}
        \depedge[edge start x offset=-1pt,edge unit distance=2.75ex]{4}{7}{acl:relcl}
        \depedge[edge start x offset=-3pt]{7}{8}{advmod}
        \depedge{11}{9}{case}
        \depedge{11}{10}{det}
        \depedge[edge unit distance=1ex,edge below,dashed]{8}{11}{obl}
        \depedge[edge start x offset=-1pt,edge unit distance=1.2ex]{2}{12}{punct}
    \end{dependency}
  \caption{\label{fig:example_relcl_ud}UD.}
\end{subfigure}}
\fbox{\begin{subfigure}{0.5\textwidth}
  \centering\scalebox{.7}{
	\begin{tikzpicture}[->,level distance=11.5mm,
	  level 1/.style={sibling distance=5cm},
	  level 2/.style={sibling distance=18mm},
	  level 3/.style={sibling distance=9mm},
	  level 4/.style={sibling distance=9mm,level distance=14mm},
	  every circle node/.append style={fill=black}]
	  \tikzstyle{word} = [font=\rmfamily,color=black]
	  \node (1_1) [circle] {}
	  {
	  child {node (1_2) [circle] {}
	    {
	    child {node {}
	    {
    	    child {node (1_6) [word] {He}  edge from parent [draw=none]}
    	} edge from parent [draw=none]}
	    child {node {}
	    {
    	    child {node (1_3) [word] {had}  edge from parent [draw=none]}
    	} edge from parent [draw=none]}
	    child {node (1_7) [circle] {}
	      {
	      child {node (1_14) [word] {a}  edge from parent node[midway,above,sloped]  {\scriptsize det}}
	      child {node (1_8) [word] {robe}  edge from parent node[midway,above,sloped]  {\scriptsize head}}
	      } edge from parent node[midway,above,sloped]  {\scriptsize obj}}
	    } edge from parent node[midway,above,sloped]  {\scriptsize head}}
	  child {node (1_4) [circle] {}
	    {
	    child {node {}
	    {
    	    child {node (1_15) [word] {that}  edge from parent [draw=none]}
    	} edge from parent [draw=none]}
	    child {node (1_5) [circle] {}
	      {
	      child {node (1_9) [word] {was}  edge from parent node[midway,above,sloped]  {\scriptsize aux}}
	      child {node (1_10) [word] {made}  edge from parent node[midway,above,sloped]  {\scriptsize head}}
	      } edge from parent node[midway,above,sloped]  {\scriptsize head}}
	    child {node (1_11) [circle] {}
	      {
	      child {node (1_12) [word] {back}  edge from parent node[midway,above,sloped]  {\scriptsize head}}
	      child {node (1_16) [circle] {}
	        {
	        child {node (1_18) [word] {in}  edge from parent node[midway,above,sloped]  {\scriptsize case}}
	        child {node (1_19) [word] {the}  edge from parent node[midway,above,sloped]  {\scriptsize det}}
	        child {node (1_17) [word] {'60s}  edge from parent node[midway,above,sloped]  {\scriptsize head}}
	        child {node (1_20) [word] {.}  edge from parent node[midway,above,sloped]  {\scriptsize punct}}
	        } edge from parent node[midway,above,sloped]  {\scriptsize obl}}
	      } edge from parent node[midway,above,sloped]  {\scriptsize advmod}}
	    } edge from parent node[midway,above,sloped]  {\scriptsize acl}}
	  };
	  \draw[->] (1_2) to node [midway,above,sloped] {\scriptsize nsubj} (1_6);
	  \draw[->] (1_2) to node [midway,above,sloped] {\scriptsize head} (1_3);
	  \draw[->] (1_4) to node [midway,above,sloped] {\scriptsize nsubj} (1_15);
	  \draw[dashed,->] (1_4) to node [midway,above,sloped] {\scriptsize nsubj} (1_7);
	  \draw[dashed,->] (1_7) to node [midway,above,sloped] {\scriptsize ref} (1_15);
	\end{tikzpicture}}
  \captionof{figure}{UD converted to unified DAG format.}\label{fig:converted_example_relcl}
\end{subfigure}}
\cprotect\caption{(a) \verb|reviews-255261-0007| (\verb|UD_English-EWT|), containing a relative clause,
and (b) the same graph after conversion to the unified DAG format.
The cycle is removed due to the non-terminal nodes introduced in the conversion.}
\label{fig:example_relcl}
\end{figure*}

\section{Unified DAG Format}\label{sec:format}

To apply TUPA to UD parsing,
we convert UD trees and graphs into a unified DAG format \cite{hershcovich2018multitask}.
The format consists of a rooted DAG, where the tokens are the terminal
nodes.\footnote{Our conversion code supports full conversion between UCCA and UD,
among other representation schemes,
and is publicly available at \url{http://github.com/danielhers/semstr/tree/master/semstr/conversion}.}
Edges are labeled (but not nodes),
and are divided into \textit{primary} and \textit{remote} edges,
where the primary edges form a tree (all nodes have at most one primary parent,
and the root has none).
Remote edges (denoted as dashed edges in Figure~\ref{fig:converted_examples})
enable reentrancy, and thus form a DAG together with primary edges.
Figure~\ref{fig:converted_examples} shows an example UCCA graph,
and a UD graph (containing two enhanced dependencies) before and after conversion.
Both annotate the same sentence from the
English Web Treebank \cite{L14-1067}\footnote{\url{https://catalog.ldc.upenn.edu/LDC2012T13}}.

\paragraph{Conversion protocol.}

To convert UD into the unified DAG format,
we add a pre-terminal for each token,
and attach the pre-terminals according to the original dependency edges:
traversing the tree from the root down, for each head token we create a non-terminal
parent with the edge label {\it head},
and add the node's dependents as children of the created non-terminal node
(see Figure~\ref{fig:converted_example_ud}).
This creates a constituency-like structure,
which is supported by TUPA's transition set (see \S\ref{sec:transition_set}).

Although the enhanced dependency graph is not necessarily a supergraph
of the basic dependency tree,
the graph we convert to the unified DAG format is their union:
any enhanced dependnecies that are distinct from the basic dependency of a
node (by having a different head or universal dependency relation)
are converted to \textit{remote edges} in the unified DAG format.

To convert graphs in the unified DAG format back into dependency graphs,
we collapse all \textit{head} edges,
determining for each terminal what is the highest non-terminal headed by it,
and then attaching the terminals to each other according to the edges among
their headed non-terminals.

\paragraph{Input format.}

Enhanced dependencies are encoded in the 9th column of the \mbox{CoNLL-U} format,
by an additional head index, followed by a colon and dependency relation.
Multiple enhanced dependencies for the same node are separated by pipes.
Figure~\ref{fig:enhanced_conllu} demonstrates this format.
Note that if the basic dependency is repeated in the enhanced graph
(\verb|3:nsubj:pass| in the example), we do not treat it as an enhanced 
dependency, so that the converted graph will only contain each edge once.
In addition to the UD relations defined in the basic
representations, enhanced dependencies may contain the relation \verb|ref|,
used for relative clauses.
In addition, they may contain more specific relation subtypes,
and optionally also case information.

\paragraph{Language-specific extensions and case information.}

Dependencies may contain language-specific relation subtypes,
encoded as a suffix separated from the universal relation by a colon.
These extensions are ignored by the parsing evaluation metrics,
so for example, the subtyped relation \verb|nsubj:pass| (Figure~\ref{fig:original_examples})
is considered the same as the universal relation \verb|nsubj|
for evaluation purposes.
In the enhanced dependencies,
these suffixes may also contain case information,
which may be represented by the lemma of an adposition.
For example, the ``peace''~$\to$~``earth'' dependency in Figure~\ref{fig:example_ellipsis}
is augmented as \verb|nmod:on| in the enhanced graph
(not shown in the figure because it shares the universal relation with the basic dependency).

In the conversion process, we strip any language-specific extensions
from both basic and enhanced dependencies,
leaving only the universal relations.
Consequently, case information that might be encoded in the enhanced
dependencies is lost, and we do not handle it in our current system.

\paragraph{Ellipsis and null nodes.}

In addition to enhanced dependencies, the enhanced UD representation
adds null nodes to represented elided predicates.
These, too, are ignored in the standard evaluation.
An example is shown in Figure~\ref{fig:example_ellipsis},
where an elided ``wish'' is represented by the node E9.1.
The elided predicate's dependents are attached to its argument ``peace''
in the basic representation, and the argument itself is attached as an \verb|orphan|.
In the enhanced representation, all arguments are attached to the null node as if
the elided predicate was present.

While UCCA supports empty nodes without surface realization
in the form of \textit{implicit units},
previous work on UCCA parsing has removed these from the graphs.
We do the same for UD parsing, dropping null nodes and their
associated dependencies upon conversion to the unified DAG format.
We leave parsing elided predicates for future work.

\paragraph{Propagation of conjuncts.}

Enhanced dependencies contain dependencies between conjoined predicates
and their arguments, and between predicates and their conjoined arguments
or modifiers.
While these relations can often be inferred from the basic dependencies,
in many cases they require semantic knowledge to parse correctly.
For example, in Figure~\ref{fig:example_conj},
the enhanced dependencies represent the shared subject (``he'') among the
conjoined predicates (``went'' and ``finished''),
and the conjoined modifiers (``efficiently'' and ``promptly'')
for the second predicate (``finished'').
However, there are no enhanced dependencies between the first predicate
and the second predicate's modifiers (e.g. ``went''~$\to$~``efficiently''),
as semantically only the subject is shared and not the modifiers.

\paragraph{Relative clauses.}

Finally, enhanced graphs attach predicates of relative clauses directly
to the antecedent modified by the relative clause, adding a \verb|ref|
dependency between the antecedent and the relative pronoun.
An example is shown in Figure~\ref{fig:example_relcl_ud}.
While these graphs may contain cycles
(``robe''~$\leftrightarrow$~``made''
in the example), they are removed upon conversion to
the unified DAG format by the introduction of non-terminal nodes
(see Figure~\ref{fig:converted_example_relcl}).

%%%%%%%%%%%%%%%%%%%%%%%%%%%%%%%%%%%%%%%%%%%%%%%%%%%%%%%%%%%%%%%
\section{General Transition-based DAG Parser}\label{sec:model}

We now turn to describing TUPA \cite{hershcovich2017a,hershcovich2018multitask},
a general transition-based parser \cite{Nivre03anefficient}.
TUPA uses an extended set of transitions and features that supports
reentrancies, discontinuities and non-terminal nodes.
The parser state is composed of a buffer $B$ of tokens and nodes to be processed,
a stack $S$ of nodes currently being processed,
and a graph $G=(V,E,\ell)$ of constructed nodes and edges,
where $V$ is the set of \emph{nodes}, $E$ is the set of \emph{edges},
and $\ell : E \to L$ is the \emph{label} function, $L$ being the set of possible labels.
Some states are marked as \textit{terminal}, meaning that $G$ is the final output.
A classifier is used at each step to select the next transition based on features
encoding the parser's current state.
During training, an oracle creates training instances for the classifier,
based on gold-standard annotations.

\begin{figure*}
	\begin{adjustbox}{width=\textwidth,margin=3pt,frame}
	\begin{tabular}{llll|l|llllc|c}
		\multicolumn{4}{c|}{\textbf{\small Before Transition}} & \textbf{\small Transition} & \multicolumn{5}{c|}{\textbf{\small After Transition}} & \textbf{\small Condition} \\
		\textbf{\footnotesize Stack} & \textbf{\footnotesize Buffer} & \textbf{\footnotesize Nodes} & \textbf{\footnotesize Edges} & & \textbf{\footnotesize Stack} & \textbf{\footnotesize Buffer} & \textbf{\footnotesize Nodes} & \textbf{\footnotesize Edges} & \textbf{\footnotesize Terminal?} & \\
		$S$ & $x \;|\; B$ & $V$ & $E$ & \textsc{Shift} & $S \;|\; x$ & $B$ & $V$ & $E$ & $-$ & \\
		$S \;|\; x$ & $B$ & $V$ & $E$ & \textsc{Reduce} & $S$ & $B$ & $V$ & $E$ & $-$ & \\
		$S \;|\; x$ & $B$ & $V$ & $E$ & \textsc{Node$_X$} & $S \;|\; x$ & $y \;|\; B$ & $V \cup \{ y \}$ & $E \cup \{ (y,x)_X \}$ & $-$ &
		$x \neq \mathrm{root}$ \\
		$S \;|\; y,x$ & $B$ & $V$ & $E$ & \textsc{Left-Edge$_X$} & $S \;|\; y,x$ & $B$ & $V$ & $E \cup \{ (x,y)_X \}$ & $-$ &
		\multirow{4}{50pt}{\vspace{-5mm}\[\left\{\begin{array}{l}
		x \not\in w_{1:n},\\
		y \neq \mathrm{root},\\
		y \not\leadsto_G x
		\end{array}\right.\]} \\
		$S \;|\; x,y$ & $B$ & $V$ & $E$ & \textsc{Right-Edge$_X$} & $S \;|\; x,y$ & $B$ & $V$ & $E \cup \{ (x,y)_X \}$ & $-$ & \\
		$S \;|\; y,x$ & $B$ & $V$ & $E$ & \textsc{Left-Remote$_X$} & $S \;|\; y,x$ & $B$ & $V$ & $E \cup \{ (x,y)_X^* \}$ & $-$ & \\
		$S \;|\; x,y$ & $B$ & $V$ & $E$ & \textsc{Right-Remote$_X$} & $S \;|\; x,y$ & $B$ & $V$ & $E \cup \{ (x,y)_X^* \}$ & $-$ & \\
		$S \;|\; x,y$ & $B$ & $V$ & $E$ & \textsc{Swap} & $S \;|\; y$ & $x \;|\; B$ & $V$ & $E$ & $-$ &
		$\mathrm{i}(x) < \mathrm{i}(y)$ \\
		$[\mathrm{root}]$ & $\emptyset$ & $V$ & $E$ & \textsc{Finish} & $\emptyset$ & $\emptyset$ & $V$ & $E$ & $+$ & \\
	\end{tabular}
	\end{adjustbox}
	\caption{\label{fig:transitions}
	  The transition set of TUPA.
	  We write the stack with its top to the right and the buffer with its head to the left.
	  $(\cdot,\cdot)_X$ denotes a primary $X$-labeled edge, and $(\cdot,\cdot)_X^*$ a remote $X$-labeled edge.
	  $\mathrm{i}(x)$ is the swap index (see \S\ref{sec:constraints}).
	  In addition to the specified conditions,
	  the prospective child in an \textsc{Edge} transition must not already have a primary parent.
	}
\end{figure*}

\subsection{Transition Set}\label{sec:transition_set}
Given a sequence of tokens $w_1, \ldots, w_n$,
we predict a rooted graph $G$ whose terminals are the tokens.
Parsing starts with the root node on the stack,
and the input tokens in the buffer.

The TUPA transition set, shown in Figure~\ref{fig:transitions}, includes
the standard \textsc{Shift} and \textsc{Reduce} operations,
\textsc{Node$_X$} for creating a new non-terminal node and an $X$-labeled edge,
\textsc{Left-Edge$_X$} and \textsc{Right-Edge$_X$} to create a new primary $X$-labeled edge,
\textsc{Left-Remote$_X$} and \textsc{Right-Remote$_X$} to create a new remote $X$-labeled edge,
\textsc{Swap} to handle discontinuous nodes,
and \textsc{Finish} to mark the state as terminal.

The \textsc{Remote$_X$} transitions are not required for parsing trees,
but as we treat the problem as general DAG parsing due to the inclusion of enhanced dependencies,
we include these transitions.

\begin{figure}[t]
   \begin{tikzpicture}[level distance=8mm, sibling distance=1cm]
   \node[anchor=west] at (0,1.5) {Parser state};
   \draw[color=gray,dashed] (0,-1.2) rectangle (7.5,1.25);
   \draw[color=gray] (.1,0) rectangle (2,.5);
   \node[anchor=west] at (.1,.8) {$S$};
   \node[fill=black, circle] at (.4,.275) {};
   \node[fill=purple, circle] at (.9,.275) {};
   \node[anchor=west] at (1.1,.275) {\scriptsize\ttfamily made};
   \draw[color=gray] (2.11,0) rectangle (5.1,.5);
   \node[anchor=west] at (2.11,.8) {$B$};
   \node[anchor=west] at (2.11,.26) {\scriptsize\ttfamily to feel very wel\ldots};
   \node[anchor=west] at (5.1,.8) {$G$};
   \node[fill=black, circle] at (6.35,.75) {}   
     child {node  {\small\ttfamily We} edge from parent [->] node[left] {\small nsubj}}
     child {node [fill=purple, circle] {}
     {
       child {node {\small\ttfamily were} edge from parent [->] node[right] {\small aux}}
     } edge from parent [->] node[right] {\small head} };
   \end{tikzpicture}
   \begin{tikzpicture}[->]
   \node[anchor=west] at (0,6) {Classifier};
   \tiny
   \tikzstyle{main}=[rounded rectangle, minimum size=7mm, draw=black!80, node distance=12mm]
   \node[main] (specific) at (3.5,3.5) {\small BiLSTM};
   \node (embeddings) at (3.5,1.7) {\small Embeddings};
   \foreach \i/\word in {0/{We},2/{were},5/{welcome},7/{.}} {
       \node (x\i) at (\i,1) {\small\ttfamily\word};
       \node[main] (e\i) at (\i,2.2) {};
       \path (x\i) edge (e\i);
       \path (e\i) edge (specific);
   }
    \node (x4) at (3.5,1) {\large\ldots};
    \node[main] (mlp) at (3.5,4.6) {\small MLP};
    \path (specific) edge (mlp);
    \coordinate (state) at (6.5,6.3);
    \path (state) edge [bend left] (mlp);
    \node (transition) at (3.5,5.8) {transition};
    \path (mlp) edge node[right] {softmax} (transition);
   \end{tikzpicture}
\caption{Illustration of the TUPA model, adapted from \citet{hershcovich2018multitask}.
Top: parser state (stack, buffer and intermediate graph).
		Bottom: BiLTSM architecture.
		Vector representation for the input tokens is computed
		by two layers of bidirectional LSTMs.
		The vectors for specific tokens are concatenated with
		embedding and numeric features from the parser state
		(for existing edge labels, number of children, etc.),
		and fed into the MLP for selecting the next transition.}\label{fig:single_model}
\end{figure}
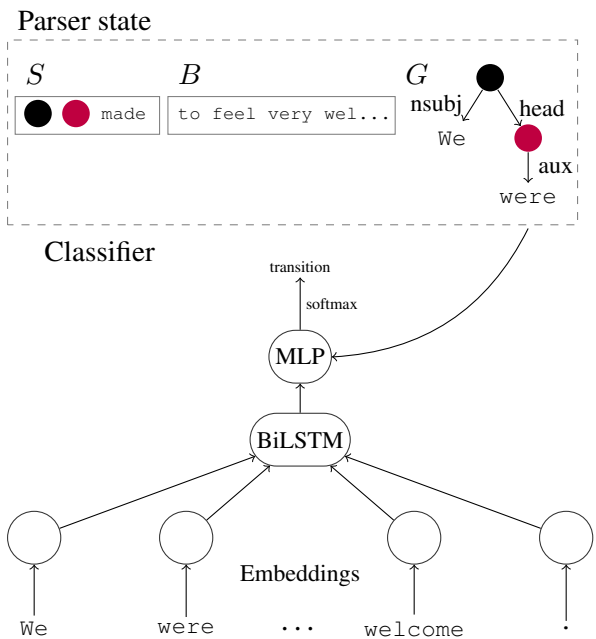

%%%%%%%%%%%%%%%%%%%%%%%%%%%%%%%%%%%%%%%%%%%%%%%%%%%%%%%%%%%%%%%%%%%%%%%%%%%%%%%%%
\subsection{Transition Classifier}\label{sec:classifier}

To predict the next transition at each step,
TUPA uses a BiLSTM with feature embeddings as inputs,
followed by an MLP and a softmax layer for classification.
The model is illustrated in Figure~\ref{fig:single_model}.
Inference is performed greedily,
and training is done with an oracle that yields the set of all optimal 
transitions at a given state (those that lead to a state from which the gold graph is still reachable).
Out of this set, the actual transition performed in training is the one
with the highest score given by the classifier,
which is trained to maximize the sum of log-likelihoods of all 
optimal transitions at each step.

\paragraph{Features.}

We use vector embeddings
representing the words, lemmas, coarse (universal) POS tags and fine-grained POS tags,
provided by UDPipe 1.2 during test.
For training, we use the gold-annotated lemmas and POS tags.
In addition, we use one-character prefix, three-character suffix,
shape (capturing orthographic features, e.g., ``Xxxx'') and named entity type,
provided by spaCy;\footnote{\url{http://spacy.io}}
punctuation and gap type features \cite{maier-lichte:2016:DiscoNLP},
and previously predicted edge labels and parser actions.
These embeddings are initialized randomly, except for the word embeddings,
which are initialized with the 250K most frequent word vectors from fastText
for each language
\cite{bojanowski2016enriching},\footnote{\url{http://fasttext.cc}}
pre-trained over Wikipedia and updated during training.
We do not use word embeddings for languages without pre-trained fastText vectors
(Ancient Greek, North Sami and Old French).

To the feature embeddings, we concatenate
numeric features representing the node height, number of (remote) parents and children,
and the ratio between the number of terminals to total number of nodes in the graph $G$.

Table~\ref{tab:features} lists all feature used for the classifier.
Numeric features are taken as they are, whereas categorical features are mapped to real-valued embedding
vectors.
For each non-terminal node,
we select a \textit{head terminal} for feature extraction,
by traversing down the graph according to
a priority order on edge labels (otherwise selecting the leftmost child).
The priority order is:
\begin{verbatim}
parataxis, conj, advcl, xcomp
\end{verbatim}

\begin{table}[h]
\centering
\small
\begin{tabular}{l|l}
\hline
\bf Nodes & \\
$s_0$ & \texttt{wmtuepT\#\^{}\$xhqyPCIEMN} \\
$s_1$ & \texttt{wmtueT\#\^{}\$xhyN} \\
$s_2$ & \texttt{wmtueT\#\^{}\$xhy} \\
$s_3$ & \texttt{wmtueT\#\^{}\$xhyN} \\
$b_0$ & \texttt{wmtuT\#\^{}\$hPCIEMN} \\
$b_1, b_2, b_3$ & \texttt{wmtuT\#\^{}\$} \\
\multirow{3}{80pt}{$s_0l, s_0r, s_1l, s_1r,$ $s_0ll, s_0lr,s_0rl, s_0rr,$ $s_1ll, s_1lr, s_1rl, s_1rr$} &
    \texttt{wme\#\^{}\$} \\\\\\
\multirow{2}{80pt}{$s_0L, s_0R, s_1L,$ $s_1R, b_0L, b_0R$} & \texttt{wme\#\^{}\$} \\\\
\hline
\bf Edges & \\
\multirow{2}{80pt}{$s_0 \to s_1, s_0 \to b_0,$ $s_1 \to s_0, b_0 \to s_0$} & \texttt{x} \\\\
$s_0 \to b_0, b_0 \to s_0$ & \texttt{e} \\
\hline
\bf Past actions \\
$a_0, a_1$ & \texttt{eA} \\
\hline
\bf Global & \texttt{node ratio}
\end{tabular}
\caption{Transition classifier features.\label{tab:features}\\
$s_i$: stack node $i$ from the top.
$b_i$: buffer node $i$.\\
$xl$, $xr$ ($xL$, $xR$): $x$'s leftmost and rightmost children (parents).
\texttt{w}: head terminal text.
\texttt{m}: lemma.
\texttt{u}: coarse (universal) POS tag.
\texttt{t}: fine-grained POS tag.
\texttt{h}: node's height.
\texttt{e}: label of its first incoming edge.
\texttt{p}: any separator punctuation between $s_0$ and $s_1$.
\texttt{q}: count of any separator punctuation between $s_0$ and $s_1$.
\texttt{x}: numeric value of gap type \cite{maier-lichte:2016:DiscoNLP}.
\texttt{y}: sum of gap lengths.
\texttt{P}, \texttt{C}, \texttt{I}, \texttt{E}, and \texttt{M}: number of
parents, children, implicit children, remote children, and remote parents.
\texttt{N}: numeric value of the head terminal's named entity IOB indicator.
\texttt{T}: named entity type.
\texttt{\#}: word shape (capturing orthographic features, e.g. "Xxxx" or "dd").
\texttt{\^{}}: one-character prefix.
\texttt{\$}: three-character suffix.\\
$x \to y$ refers to the existing edge from $x$ to $y$.
\texttt{x} is an indicator feature, taking the value of 1 if the edge exists or 0 otherwise,
\texttt{e} refers to the edge label, and
$a_i$ to the transition taken $i+1$ steps ago.\\
\texttt{A} refers to the action type (e.g. \textsc{shift}/\textsc{right-edge}/\textsc{node}), and
\texttt{e} to the edge label created by the action.\\
\texttt{node ratio} is the ratio between non-terminals and terminals \cite{hershcovich2017a}.}
\end{table}

\subsection{Constraints}\label{sec:constraints}
During training and parsing, we apply constraints on the parser state
to limit the possible transitions to valid ones.

A generic constraint implemented in TUPA is that stack nodes 
that have been swapped
should not be swapped again \cite{hershcovich2018multitask}.
 To implement this constraint, we define a \textit{swap index}
 for each node, assigned when the node is created.
 At initialization, only the root node and terminals exist.
 We assign the root a swap index of 0, and for each terminal, its
 position in the text (starting at 1).
 Whenever a node is created as a result of a \textsc{Node}
 transition, its swap index is the arithmetic
 mean of the swap indices of the stack top and buffer head.
 
In addition, we enforce UD-specific constraints, resulting from
the nature of the converted DAG format:
every non-terminal node must have a single outgoing \verb|head| edge:
once it has one, it may not get another, and
until it does, the node may not be reduced.

\section{Training details}\label{sec:details}

The model is implemented using DyNet v2.0.3
\cite{neubig2017dynet}.\footnote{\url{http://dynet.io}}
Unless otherwise noted, we use the default values provided by the package.
We use the same hyperparameters as used in previous experiments on UCCA
parsing \cite{hershcovich2018multitask},
without any hyperparameter tuning on UD treebanks.

\begin{table}[h]
\centering
\begin{tabular}{l|c|ccccc}
\hline
\bf Hyperparameter &  \bf Value \\
\hline
Pre-trained word dim. & 300 \\
Lemma dim. & 200 \\
Coarse (universal) POS tag dim. & 20 \\
Fine-grained POS tag dim. & 20 \\
Named entity dim. & 3 \\
Punctuation dim. & 1 \\
Shape dim. & 3 \\
Prefix dim. & 2 \\
Suffix dim. & 3 \\
Action dim. & 3 \\
Edge label dim. & 20 \\
\hline
MLP layers & 2 \\
MLP dimensions & 50 \\
BiLSTM layers & 2 & \\
BiLSTM dimensions & 500
\end{tabular}
\caption{Hyperparameter settings.\label{tab:hyperparams}}
\end{table}

\subsection{Hyperparameters}

We use dropout \cite{srivastava2014dropout} between MLP layers, and recurrent dropout
\cite{NIPS2016_6241} between BiLSTM layers, both with $p=0.4$.
We also use word, lemma, coarse- and fine-grained POS tag dropout
with $\alpha=0.2$
\cite{kiperwasser2016simple}: in training, the embedding for a feature value
$w$ is replaced with a zero vector with a probability of
$\frac{\alpha}{\#(w)+\alpha}$,
where $\#(w)$ is the number of occurrences of $w$ observed.
In addition, we use \textit{node dropout} \cite{hershcovich2018multitask}:
with a probability of 0.1 at each step, all features associated with a single
node in the parser state are replaced with zero vectors.
For optimization we use a minibatch size of 100, decaying all weights by $10^{-5}$ at each update,
and train with stochastic gradient descent for $50$ epochs with a learning
rate of 0.1, followed by AMSGrad \cite{j.2018on} for $250$ epochs with
$\alpha=0.001,\beta_1=0.9$ and $\beta_2=0.999$.
We found this training strategy better than using only one of the optimization methods,
similar to findings by \citet{keskar2017improving}.
We select the epoch with the best LAS-F1 on the
development set.
Other hyperparameter settings are listed in Table~\ref{tab:hyperparams}.

\subsection{Small Treebanks}

For corpora with less than 100 training sentences,
we use $750$ epochs of AMSGrad instead of $250$.
For corpora with no development set,
we use 10-fold cross-validation on the training set,
each time splitting it to 80\% training, 10\% development and 10\% validation.
We perform the normal training procedure on the training and development
subsets, and then select the model from the fold with the best LAS-F1
on the corresponding validation set.

\subsection{Multilingual Model}

For the purpose of parsing languages with no training data,
we use a delexicalized multilingual model, trained on the shuffled training sets
from all corpora, with no word, lemma, fine-grained tag, prefix and suffix features.
We train this model for two epochs using stochastic gradient descent
with a learning rate of $0.1$
(we only trained this many epochs due to time constraints).

\subsection{Out-of-domain Evaluation}

For test treebanks without corresponding training data,
but with training data in the same language, during testing
we use the model trained on the largest training treebank in the same language.

\section{Results}\label{sec:results}

Official evaluation was done on the TIRA online platform \cite{tira}.
Our system (named ``HUJI'') ranked 24th in the LAS-F1 ranking
(with an average of 53.69 over all test treebanks),
23rd by MLAS (average of 44.6) and 21st by BLEX (average of 48.05).
Since our system only performs dependency parsing and not other pipeline tasks,
we henceforth focus on LAS-F1 \cite{nivre17udw} for evaluation.

After the official evaluation period ended,
we discovered several bugs in the conversion between the CoNLL-U format
and the unified DAG format, which is used by TUPA for training and is output by it
(see~\S\ref{sec:format}).
We did not re-train TUPA on the training treebanks after fixing these bugs,
but we did re-evaluate the already trained models on all test treebanks,
and used the fixed code for converting their output to CoNLL-U.
This yielded an unofficial average test LAS-F1 of 58.48,
an improvement of 4.79 points over the official average score.
In particular, for two test sets, \verb|ar_padt| and \verb|gl_ctg|, TUPA got
a zero score in the official evaluation due to a bug with the treatment of multi-token words.
These went up to 61.9 and 71.42, respectively.
We also evaluated the trained TUPA models on all available development treebanks
after fixing the bugs.

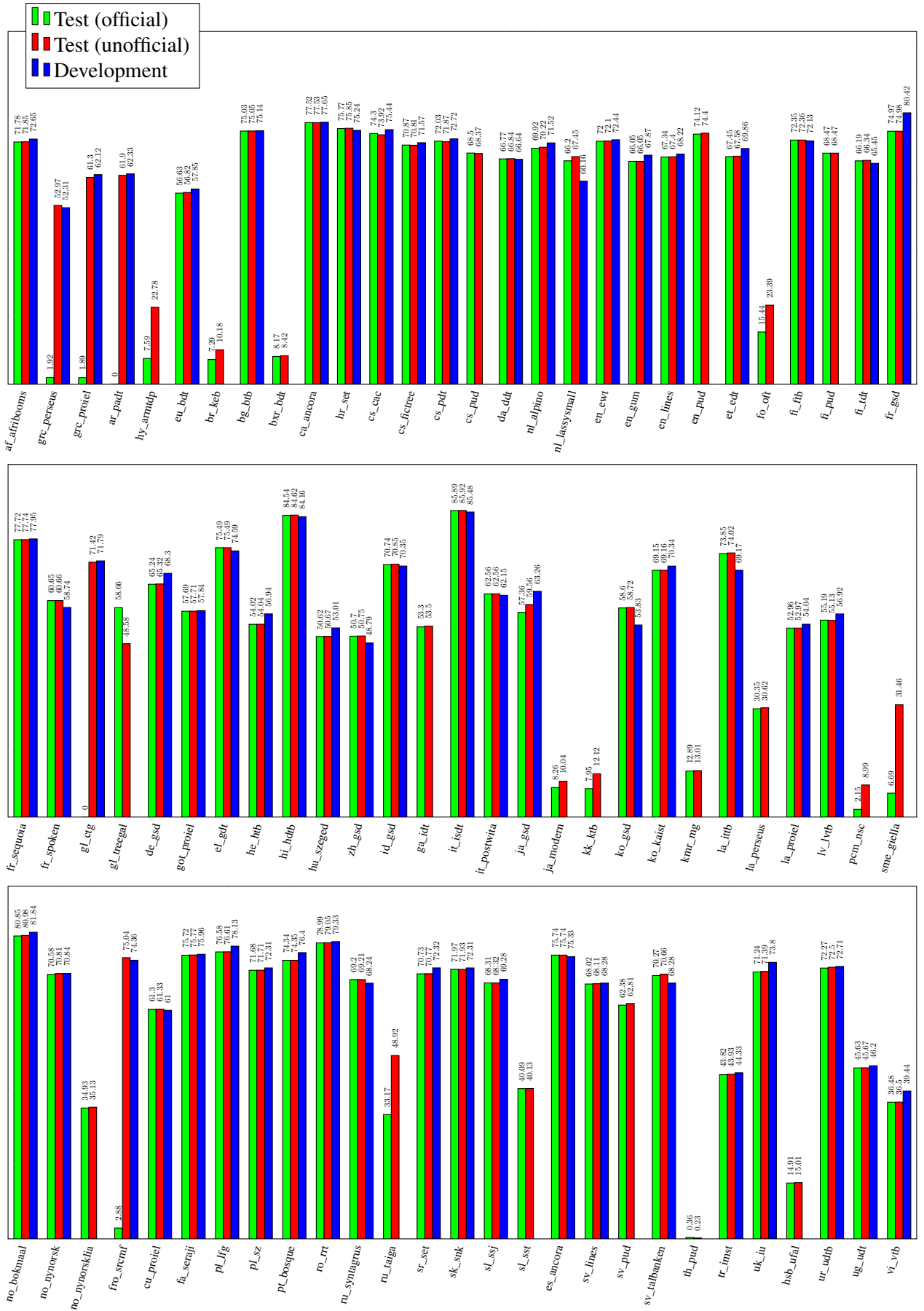
\begin{figure*}[h!]
    \begin{tikzpicture}
    \begin{axis}[
    ybar=0pt,  
    enlarge x limits={0.02},
    enlarge y limits={value=0.3,upper},
    ymin=0,
    width=18cm,
    height=8cm,
    bar width=4pt,
    xtick=data,
    xticklabels from table={\corpusa}{corpus},
    xticklabel style={font=\tiny,rotate=75,anchor=east},
    xtick align=inside,
    xticklabel pos=left,
    yticklabels=none,
    tickwidth=0pt,
    legend style={at={(axis cs:0,113)},anchor=north west},
    legend cell align={left},
    nodes near coords={
     \scalebox{.4}{\pgfmathprintnumber[precision=2]{\pgfplotspointmeta}}
    },
    every node near coord/.append style={rotate=90,anchor=west}
    ]
    \addplot[select coords between index={0}{27},fill=green]table[x expr=\coordindex,meta=corpus,y=official]{udst2018scores.txt};
    \addplot[select coords between index={0}{27},fill=red]table[x expr=\coordindex,meta=corpus,y=test]{udst2018scores.txt};
    \addplot[select coords between index={0}{27},fill=blue]table[x expr=\coordindex,meta=corpus,y=dev]{udst2018scores.txt};
    \legend{Test (official), Test (unofficial), Development}
    \end{axis}
    \end{tikzpicture}
   
    \begin{tikzpicture}
    \begin{axis}[
    ybar=0pt,  
    enlarge x limits={0.02},
    enlarge y limits={value=0.15,upper},
    ymin=0,
    width=18cm,
    height=8cm,
    bar width=4pt,
    xtick=data,
    xticklabels from table={\corpusb}{corpus},
    xticklabel style={font=\tiny,rotate=75,anchor=east},
    xtick align=inside,
    xticklabel pos=left,
    yticklabels=none,
    tickwidth=0pt,
    nodes near coords={
     \scalebox{.4}{\pgfmathprintnumber[precision=2]{\pgfplotspointmeta}}
    },
    every node near coord/.append style={rotate=90, anchor=west}
    ]
    \addplot[select coords between index={28}{54},fill=green]table[x expr=\coordindex,meta=corpus,y=official]{udst2018scores.txt};
    \addplot[select coords between index={28}{54},fill=red]table[x expr=\coordindex,meta=corpus,y=test]{udst2018scores.txt};
    \addplot[select coords between index={28}{54},fill=blue]table[x expr=\coordindex,meta=corpus,y=dev]{udst2018scores.txt};
    \end{axis}
    \end{tikzpicture}
   
    \begin{tikzpicture}
    \begin{axis}[
    ybar=0pt,  
    enlarge x limits={0.02},
    enlarge y limits={value=0.15,upper},
    ymin=0,
    width=18cm,
    height=8cm,
    bar width=4pt,
    xtick=data,
    xticklabels from table={\corpusc}{corpus},
    xticklabel style={font=\tiny,rotate=75,anchor=east},
    xtick align=inside,
    xticklabel pos=left,
    yticklabels=none,
    tickwidth=0pt,
    nodes near coords={
     \scalebox{.4}{\pgfmathprintnumber[precision=2]{\pgfplotspointmeta}}
    },
    every node near coord/.append style={rotate=90, anchor=west}
    ]
    \addplot[select coords between index={55}{81},fill=green]table[x expr=\coordindex,meta=corpus,y=official]{udst2018scores.txt};
    \addplot[select coords between index={55}{81},fill=red]table[x expr=\coordindex,meta=corpus,y=test]{udst2018scores.txt};
    \addplot[select coords between index={55}{81},fill=blue]table[x expr=\coordindex,meta=corpus,y=dev]{udst2018scores.txt};
    \end{axis}
    \end{tikzpicture}
    \caption{TUPA's LAS-F1 per treebank: official and unofficial test scores, and development scores (where available).
    \label{fig:test_per_corpus}}
\end{figure*}

Table~\ref{tab:overall_results} presents the
averaged scores on the shared task test sets,
and Figure~\ref{fig:test_per_corpus} the (official and unofficial) LAS-F1
scores obtained by TUPA on each of the test and development treebanks.

\begin{table}
\begin{tabular}{lccc}
\hline
& \multirow{2}{12mm}{\bf TUPA {\small(official)}} & \multirow{2}{15mm}{\bf TUPA {\small(unofficial)}}
& \multirow{2}{14mm}{\bf UDPipe {\small(baseline)}} \\\\
\hline
All treebanks & 53.69 & 58.48 & 65.80 \\
Big treebanks & 62.07 & 67.36 & 74.14 \\
PUD treebanks & 56.35 & 56.82 & 66.63 \\
Small treebanks & 36.74 & 41.19 & 55.01 \\
Low-resource & 8.53 & 12.68 & 17.17
\end{tabular}
\caption{Aggregated test LAS-F1 scores
for our system (TUPA) and the baseline system (UDPipe 1.2).
\label{tab:overall_results}}
\end{table}

\catcode`\_=12
\pgfplotstableread{
corpus
af_afribooms
grc_perseus
grc_proiel
ar_padt
hy_armtdp
eu_bdt
br_keb
bg_btb
bxr_bdt
ca_ancora
hr_set
cs_cac
cs_fictree
cs_pdt
cs_pud
da_ddt
nl_alpino
nl_lassysmall
en_ewt
en_gum
en_lines
en_pud
et_edt
fo_oft
fi_ftb
fi_pud
fi_tdt
fr_gsd
}\corpusa
\pgfplotstableread{
corpus
fr_sequoia
fr_spoken
gl_ctg
gl_treegal
de_gsd
got_proiel
el_gdt
he_htb
hi_hdtb
hu_szeged
zh_gsd
id_gsd
ga_idt
it_isdt
it_postwita
ja_gsd
ja_modern
kk_ktb
ko_gsd
ko_kaist
kmr_mg
la_ittb
la_perseus
la_proiel
lv_lvtb
pcm_nsc
sme_giella
}\corpusb
\pgfplotstableread{
corpus
no_bokmaal
no_nynorsk
no_nynorsklia
fro_srcmf
cu_proiel
fa_seraji
pl_lfg
pl_sz
pt_bosque
ro_rrt
ru_syntagrus
ru_taiga
sr_set
sk_snk
sl_ssj
sl_sst
es_ancora
sv_lines
sv_pud
sv_talbanken
th_pud
tr_imst
uk_iu
hsb_ufal
ur_udtb
ug_udt
vi_vtb
}\corpusc

\catcode`\_=8

\subsection{Evaluation on Enhanced Dependencies}\label{sec:enhanced_results}

Since the official evaluation ignores enhanced dependencies,
we evaluate them separately using a modified version of the shared task evaluation
script\footnote{\url{https://github.com/CoNLL-UD-2018/HUJI/blob/master/tupa/scripts/conll18_ud_eval.py}}.
We calculate the \textit{enhanced LAS},
identical to the standard LAS except that the set of dependencies
in both gold and predicted graphs are the enhanced dependencies instead
of the basic dependencies:
ignoring null nodes and any enhanced dependency sharing a head with a basic one,
we align the words in the gold graph and the system's graph as in the standard LAS,
and define
\[
P=\frac{\#\mathrm{correct}}{\#\mathrm{system}},
R=\frac{\#\mathrm{correct}}{\#\mathrm{gold}},
F1=2\cdot\frac{P\cdot R}{P+R}.
\]

Table~\ref{tab:enhanced} lists the enhanced LAS precision, recall and F1 score
on the test treebanks with any enhanced dependencies,
as well as the percentage of enhanced dependencies in each test treebank,
calculated as
$100 \cdot \frac{\#\mathrm{enhanced}}{\#\mathrm{enhanced} + \#\mathrm{words}}$.

Just as remote edges in UCCA parsing are more challenging
than primary edges \cite{hershcovich2017a},
parsing enhanced dependencies is a harder task than standard UD parsing,
as the scores demonstrate.
However, TUPA learns them successfully, getting as much as 56.55 enhanced LAS-F1
(on the Polish LFG test set).

\catcode`\_=12
\begin{table}[t]
\begin{tabular}{l|ccc|c}
\hline
& \multicolumn{3}{c|}{Enhanced LAS} & \multirow{2}{11mm}{\% Enhanced} \\
\bf Treebank & \bf P & \bf R & \bf F1 & \\
\hline
ar_padt & 28.51 & 16.24 & 20.69 & 5.30\\ 
cs_cac & 54.94 & 35.69 & 43.27 & 7.57\\ 
cs_fictree & 48.78 & 18.53 & 26.85 & 4.30\\ 
cs_pdt & 49.46 & 26.47 & 34.48 & 4.61\\ 
nl_alpino & 56.04 & 50.81 & 53.30 & 4.80\\ 
nl_lassysmall & 49.71 & 51.30 & 50.49 & 4.13\\ 
en_ewt & 57.36 & 52.05 & 54.58 & 4.36\\ 
en_pud & 58.99 & 50.00 & 54.13 & 5.14\\ 
fi_tdt & 40.20 & 31.37 & 35.24 & 7.34\\ 
lv_lvtb & 31.76 & 18.70 & 23.54 & 4.12\\ 
pl_lfg & 59.19 & 54.13 & 56.55 & 2.61\\ 
sk_snk & 37.28 & 21.61 & 27.36 & 3.91\\ 
sv_pud & 45.40 & 39.58 & 42.29 & 6.36\\ 
sv_talbanken & 50.15 & 43.20 & 46.42 & 6.89
\end{tabular}
\caption{TUPA's enhanced LAS precision, recall and F1 per test treebank with 
any enhanced dependencies,
and percentage of enhanced dependencies in test treebank.
\label{tab:enhanced}}
\end{table}
\catcode`\_=8

\subsection{Ablation Experiments}\label{sec:ablation}

The TUPA transition classifier for some of the languages uses
named entity features calculated by
spaCy.\footnote{\url{https://spacy.io/api/annotation}}
For German, Spanish, Portuguese, French, Italian, Dutch and Russian,
the spaCy named entity recognizer was trained
on Wikipedia \cite{nothman2013learning}.
However, the English model was trained on
OntoNotes\footnote{\url{https://catalog.ldc.upenn.edu/LDC2013T19}},
which is in fact not among the additional resources allowed by the shared task
organizers.
To get a fair evaluation
and to quantify the contribution of the NER features,
we re-trained TUPA on the English EWT (\verb|en_ewt|) training set
with the same hyperparameters as in our submitted model,
just without these features.
As Table~\ref{tab:ablation} shows,
removing the NER features ($-$NER) only slightly hurts the performance,
by 0.28 LAS-F1 points on the test treebank,
and 0.63 on the development treebank. 

As further ablation experiments, we tried removing POS features,
pre-trained word embeddings,
and remote edges (the latter enabling TUPA to parse enhanced dependencies).
Removing the POS features does hurt performance to a larger degree,
by 2.87 LAS-F1 points on the test set,
while removing the pre-trained word embeddings
even slightly improves the performance.
Removing remote edges and transitions from TUPA causes a very small decrease
in LAS-F1, and of course enhanced dependencies can then no longer be produced at all.

\begin{table}[t]
\begin{tabular}{l|cc|cc}
\hline
& \multicolumn{2}{c|}{\small LAS-F1} & \multicolumn{2}{c}{\small Enhanced LAS-F1} \\
\bf Model & \bf Test & \bf Dev & \bf Test & \bf Dev \\
\hline
Original & 72.10 & 72.44 & 54.58 & 57.13 \\
%Re-trained & 72.07 & 72.08 & 54.71 & 55.97\\ 
$-$NER & 71.82 & 71.81 & 55.31 & 54.65\\ 
$-$POS & 69.23 & 69.54 & 53.78 & 49.12\\ 
$-$Embed. & 72.33 & 72.55 & 56.26 & 54.54\\ 
$-$Remote & 72.08 & 72.32 & 0.00 & 0.00\\ 
%$-$Remote$-$NER & 71.36 & 71.78 & 0.00 & 0.00\\ 
%$-$Remote$-$Embed. & 72.22 & 72.44 & 0.00 & 0.00
\end{tabular}
\caption{Ablation LAS-F1 and Enhanced LAS-F1 on the English EWT development
and test set.
NER: named entity features.
POS: part-of-speech tag features (both universal and fine-grained).
Embed.: external pre-trained word embeddings (fastText).
Remote: remote edges and transitions in TUPA.
\label{tab:ablation}}
\end{table}

\section{Conclusion}\label{sec:conclusion}

We have presented the HUJI submission to the CoNLL 2018 shared task on parsing Universal Dependencies, based on TUPA, a general transition-based DAG parser.
Using a simple conversion protocol to convert UD into a unified DAG format,
training TUPA as-is on the UD treebanks yields results
close to the UDPipe baseline for most treebanks in the standard evaluation.
While other systems ignore enhanced dependencies,
TUPA learns to produce them too as part of the general dependency parsing process.
We believe that with hyperparameter tuning and more careful handling of
cross-lingual and cross-domain parsing, TUPA can be competitive on the standard metrics too.

Furthermore, the generic nature of our parser, which supports many representation schemes, as well as domains and languages,
will allow improving performance by multitask learning \citep[cf.][]{hershcovich2018multitask},
which we plan to explore in future work.

\section*{Acknowledgments}

This work was supported by the Israel Science Foundation (grant no. 929/17) and
by the HUJI Cyber Security Research Center
in conjunction with the Israel National Cyber Bureau in the Prime Minister's Office.

\bibliography{references}
\bibliographystyle{acl_natbib}

\end{document}